# 3D Whole Brain Segmentation using Spatially Localized Atlas Network Tiles

Yuankai Huo*, Zhoubing Xu, Yunxi Xiong, Katherine Aboud, Prasanna Parvathaneni, Shunxing Bao, Camilo Bermudez, Susan M. Resnick, Laurie E. Cutting, and Bennett A. Landman

*Abstract*— Detailed whole brain segmentation is an essential quantitative technique in medical image analysis, which provides a non-invasive way of measuring brain regions from a clinical acquired structural magnetic resonance imaging (MRI). Recently, deep convolution neural network (CNN) has been applied to whole brain segmentation. However, restricted by current GPU memory, 2D based methods, downsampling based 3D CNN methods, and patch-based high-resolution 3D CNN methods have been the de facto standard solutions. 3D patch-based high resolution methods typically yield superior performance among CNN approaches on detailed whole brain segmentation (>100 labels), however, whose performance are still commonly inferior compared with state-of-the-art multi-atlas segmentation methods (MAS) due to the following challenges: (1) a single network is typically used to learn both spatial and contextual information for the patches, (2) limited manually traced whole brain volumes are available (typically less than 50) for training a network. In this work, we propose the spatially localized atlas network tiles (SLANT) method to distribute multiple independent 3D fully convolutional networks (FCN) for high-resolution whole brain segmentation. To address the first challenge, multiple spatially distributed networks were used in the SLANT method, in which each network learned contextual information for a fixed spatial location. To address the second challenge, auxiliary labels on 5111 initially unlabeled scans were created by multi-atlas segmentation for training. Since the method integrated multiple traditional medical image processing methods with deep learning, we developed a containerized pipeline to deploy the end-to-end solution. From the results, the proposed method achieved superior performance compared with multi-atlas segmentation methods, while reducing the computational time from >30 hours to 15 minutes. The method has been made available in (https://github.com/MASILab/SLANTbrainSeg).

## I. Introduction

Whole brain segmentation is essential in the scientific and clinical investigation for understanding the human brain quantitatively, which provides a non-invasive tool to quantify brain structures from a single clinical acquired structural magnetic resonance imaging (MRI). A manual delineation on brain structures has been regarded as the long-held "gold standard". Yet, manual delineation is resource and time intensive, which is impractical to be deployed on a large-scale. Therefore, fully-automated algorithms have been desired to alleviate the manual efforts. In the 1990s, fuzzy c-mean methods had been used to parcellate a brain MRI to three tissues: gray matter (GM), white matter (WM), and cerebrospinal fluid (CSF) [1]. Since then, advanced whole brain segmentation methods have been proposed including, but not limited to, region growing, clustering, deformation models, and atlas-based methods [2].

Atlas-based segmentation is one of the most prominent families among the segmentation methods, which assigns tissue labels to the unlabeled images using structural MR scans as well as the corresponding manual segmentation. In atlas-based segmentation models, the deformable registration methods are typically used to spatially transfer an existing dataset (atlas) to a previously unseen target image [3-5]. Single-atlas segmentation has been successfully applied to some applications [3-5]. However, the single-atlas segmentation suffers inferior performance when targeting large inter-subject variation on anatomy [6], as reviewed in [7]. More recent approaches employ a multi-atlas paradigm as the de facto standard atlas-based segmentation framework [8, 9]. In multi-atlas segmentation, the typical framework is: (1) a set of labeled atlases are registered to a target image [10-13], and (2) the resulting label conflicts are addressed using label fusion [9, 14-25]. With intensity harmonization [26], whole brain registrations [27], and multi-atlas label fusion (MALF), state-of-the-art MAS methods are able to segment brain from a clinical acquired T1-weighted (T1w) MRI volume to more than 100 labels using a small number of manually traced and representative scans. MAS approaches have been regarded as the de facto standard whole brain segmentation methods due to their superior performance on accuracy and reproductivity.

To deal with the local anatomical variations from imperfect registrations, the patch-based methods [16, 28-32] have been proposed. Meanwhile, the multi-atlas label fusion theory has been developed to model the spatial relationships between atlases and targets in 3D patches. To improve the performance of MAS for longitudinal data, the 4D patch MALF method was proposed [33] to incorporate the probabilistic model of temporal performance of atlases to the voting-based fusion.

*Y. Huo is with the Department of Electrical Engineering and Computer Science, Vanderbilt University, Nashville, TN 37235 USA (e-mail: yuiankai.huo@vanderbilt.edu).

Z. Xu, Y. Xiong, P. Parvathaneni, S. Bao, and B. A. Landman are with the Department of Electrical Engineering and Computer Science, Vanderbilt University, TN 37235 USA

K. Aboud, and L. E. Cutting are with the Department of Special Education, Vanderbilt University, TN 37235 USA

C. Bermudez is with the Department of Biomedical Engineering, Vanderbilt University, TN 37235 USA

S. M. Resnick is with the Laboratory of Behavioral Neuroscience, National Institute on Aging, Baltimore, MD 20892 USA

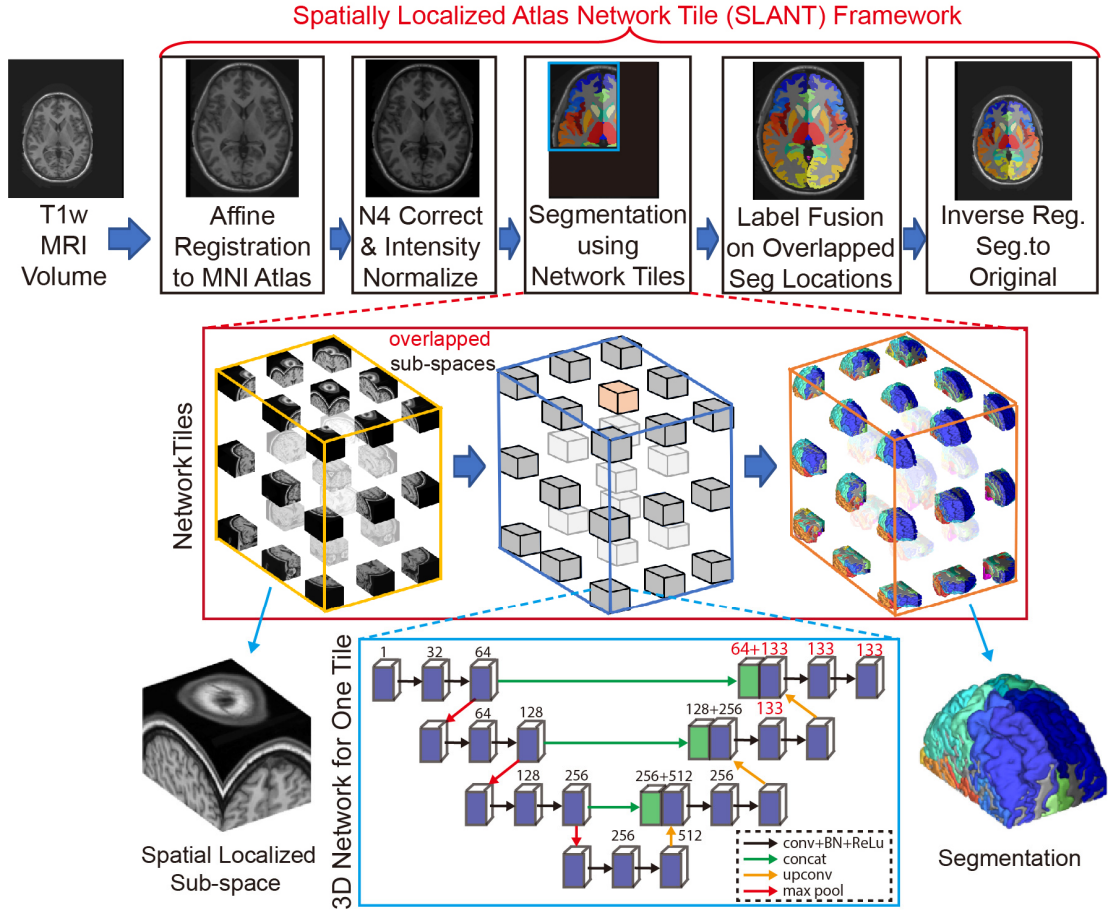

Figure 1. The proposed SLANT-27 (27 network tiles) whole brain segmentation method is presented, which combines canonical medical image processing (registration, harmonization, label fusion) with 3D network tiles. 3D U-Net framework is used as each tile, whose deconvolutional channel numbers are increased to 133. The tiles are spatially overlapped in MNI space, whose intensity inputs and segmentation outputs for one tile are visualized.

However, one of the major limitations of the traditional MAS methods is the high computational cost. Therefore, the MALF is typically performed with a small number of atlases (e.g., <100 atlases for whole brain segmentation). To utilize the larger number of atlases or even previously unlabeled scans, many previous efforts have been proposed to develop faster and more robust MAS segmentation methods using unlabeled or automatically labeled data [34, 35]. With the similar intuition, the machine learning techniques have been incorporated into MAS to replace the traditional voting based or statistical model patch MALF. A number of machine learning MALF methods have been developed to learn MAS model from the larger number of atlases [36-41]. The core idea of learning from training image patches contributes to and inspires a large number of deep convolutional neural networks (CNN) methods, including our proposed method.

Recently, CNN methods have been widely developed to applied to whole brain segmentation. The straightforward strategy of performing whole brain segmentation is to fit all brain volume to a 3D CNN based segmentation network, like U-Net [42] or V-Net [43]. Unfortunately, it is impractical to fit the clinical used high-resolution MRI (e.g., 1mm or even higher isotropic voxel size) to state-of-the-art 3D fully convolutional networks (FCN) due to the memory limitation of prevalent GPU. Another challenge of using CNN methods is that the manually traced whole brain MRI scans with detailed annotations (e.g., >100 labels) are rare commodities for any individual lab. To address the challenges of GPU memory restriction and limited training data, many previous efforts have been made. As a pioneer, de Brébisson [44] proposed a unified CNN network to learn 2D and 3D patches as well as their spatial coordinates for whole brain segmentation. Then, such network has been extended to BrainSegNet [45], which employed 2.5D patches for training a CNN network. Recently, DeepNAT [46] was proposed to perform hierarchical multi-task learning on 3D patches. These methods modeled the whole brain segmentation as a per-voxel segmentation problem. More recently, from another "image-to-image" perspective, the powerful fully convolution networks (FCN) have introduced to the whole brain segmentation. Roy et al., [47] developed a 2D based method to train an FCN network using large-scale auxiliary labels on initially unlabeled data. Although the training was in a 2D manner, Roy et al., revealed a promising direction on how to leverage the whole brain segmentation network not only using manually traced images but also using initially unlabeled data. However, 2D based segmentation methods typically yield inferior spatial consistency on the third dimension. Therefore, it is appealing to perform 3D FCN (e.g., 3D U-Net [42]) on whole brain segmentation for higher spatial consistency. Recently, multi-task learning [48], complementary learning [49], improved loss function [50], and semi-supervised learning [51] have been applied to whole brain segmentation. Among such works, Rajchl et al. proposed a novel 3D multi-task learning network structure, which was able to achieve decent segmentation performance. However, the input image size of such network is limited to $128 \times 128 \times 128$ (2,097,152 voxels) due to the GPU memory, which missed about two third of spatial information compared with 1mm isotropic brain volume in MNI space $172 \times 220 \times 156$ (5,903,040 voxels). Therefore, directly applying 3D FCN to whole brain segmentation (e.g., with 1mm isotropic resolution in MNI space or even higher resolution) is still restricted by the current graphics processing unit (GPU) memory. To address such challenges, Li et al., [52] proposed a 3D CNN based sliding window based method, which used a single network to learn the patches at the different spatial location. In such a design, the single network implicitly learns two things: (1) "where the patch is located in the brain?", and (2) "how to assign labels for that patch". (1) decides the candidate labels for such patch among 133 labels, while (2) decides how to label each voxel in the patch.

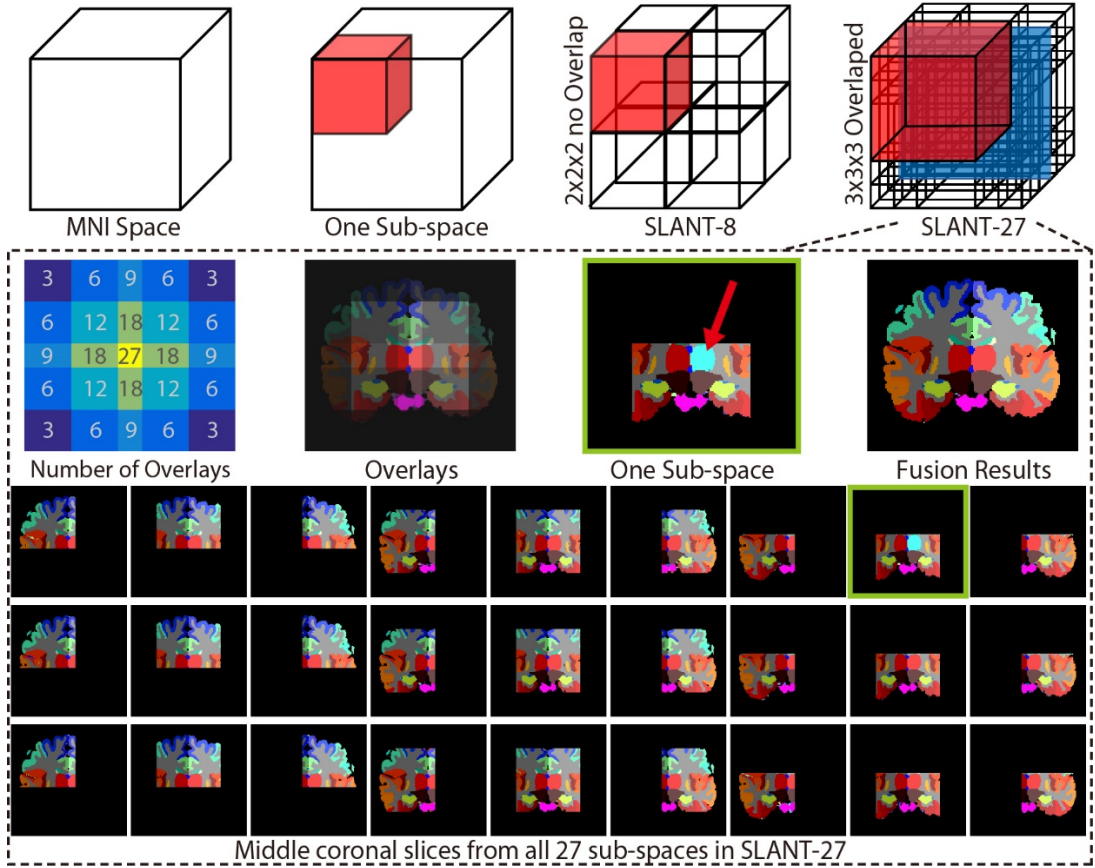

Figure 2. This figure presents the SLANT-8 and SLANT-27. SLANT-8 covered eight non-overlapped sub-spaces in MNI, while SLANT-27 covered 27 overlapped sub-spaces in MNI. Middle coronal slices from all 27 sub-spaces were visualized (lower panel). The number of overlays as well as sub-spaces' overlays were showed (middle panels). The incorrect labels (red arrow) in one sub-space were corrected in final segmentation by performing majority vote label fusion.

In this work, we propose the spatially localized atlas network tiles (SLANT) method to alleviate the difficulties of patch-based learning by employing multiple independent 3D FCN networks that each network is only responsible for a particular spatial location. As multiple networks are used, the task of each network is simplified to focus on the patches from the similar parts of the brains with smaller spatial variations (e.g., each network deals with a particular sub-section of the brain as Figure 1). To enable such strategy, affine registration and standard brain space in traditional medical image processing are used to roughly normalize each brain to the same space. Finally, the label fusion technique is used to achieve the final segmentation from the network tiles. This work extended our conference publication [53] with (1) more complete illustrations on method and experiments, (2) new surface-based validation, (3) docker implementation, (4) new baseline methods, and (5) detailed regions of interests (ROIs) level analyses. Both the docker and the source code of SLANT have been made freely available online (https://github.com/MASILab/SLANTbrainSeg).

## II. METHODS

The processing pipeline of SLANT method has been presented in Figure 1. The SLANT pipeline employed the historical efforts in medical image processing including intensity harmonization, registration, and label fusion.

*A. Registration and Intensity Harmonization*

The input of the SLANT pipeline is a single MRI T1w 3D brain scan. 45 manually traced MRI T1w 3D brain scans had been employed as the training data for training a deep convolutional network, multi-atlas segmentation, and intensity normalization. The first step in SLANT pipeline was an Affine registration from the target image to the MNI305 template [54] using NiftyReg [11]. Then, an N4 bias field correction [26] was performed to alleviate the bias from the imaging procedure. Note that MRI is a non-scaled imaging technique, which means the intensities of acquired scans varies across different scanners, and even different scans from the same scanner. Therefore, to further normalize the intensities across different scans, a regression-based intensity normalization method was introduced in the SLANT pipeline.

First, we define an MRI volume as a vector $I \in \mathbb{R}^{N \times 1}$, where $N$ is the number of voxels and $\mathbb{R}$ means real numbers. Next, $I$ was normalized to $I'$ by subtracting the mean intensity value and divided by standard deviation (std). The intensities were then harmonized by a pretrained regression model from all training atlases on "sorted intensity" in MNI 305 space. Briefly, the sorted intensity $V_s$ for $I$ was calculated by $V_s = \text{sort}(I'(mask > 0))$, where the "sort" operation rearrange the intensities from largest to smallest. The "$mask$" was a binary hard mask learned from the average brain tissue probabilistic map (averaging brain tissue label in 45 atlases) by thresholding the probability with 0.5. The $mask$ is able to exclude the non-brain tissue intensities when performing regression. To train the robust regression [55], mean sorted intensity vector $\overline{V_s}$ was obtained by averaging all $V_s$ from all atlases. When segmenting a new testing scan, we modeled the relationship between $\overline{V_s}$ (precalculated from all atlases) and the linear sort intensity vectors $V_s$ (from the testing scan) as $\overline{V_s} = \beta_1 \cdot V_s' + \beta_0$. For the testing scan, the $\beta_0$ and $\beta_1$ are learned adaptively from deploying a robust regression using "robustfit" function in Matlab, whose weight function is "huber". The $\overline{V_s}$ is learned from all atlases before deploying the segmentation, while the $V_s'$ is obtained from the testing scan when running the segmentation. After getting the coefficients $\beta_1$ and $\beta_0$, we can get normalized volume $\hat{I'}$ from sorted volume $I'$ for the testing scan as $\hat{I'} = \beta_1 \cdot I' + \beta_0$. Then, the normalized $\hat{I'}$ for the testing scan was used the next learning stage.

*B. Network Tiles*

From registration and normalization, all training and testing brain volumes were mapped to the same MNI 305 standard space, whose resolution is 1 mm isotropic with $172 \times 220 \times 156$ voxels. Since the high-resolution imaging volume could not be fitted into GPU memory using prevalent FCN networks, we employed $k$ independent 3D U-Net as a network tiles to cover the entire MNI space. Each 3D U-Net was a sub-network, whose resolution is a compromise between memory limitations and spatial resolution. For each 3D U-Net, we modified the decoder part upon the original 3D U-Net implementation to be compatible with 133 labels output. As shown in Figure 1, 133 3D output channels have been employed in the deconvolutional layers in each 3D U-Net. $j$th sub-network covers the sub-space $\psi_n$, which was presented by the corner coordinate $(x_j, y_j, z_j)$ as well as the sub-space's size $(d_x, d_y, d_z)$, $j \in \{1,2,...,k\}$

$$\psi_n = [x_n:(x_n + d_x), y_n:(y_n + d_y), z_n:(z_n + d_z)] \quad (1)$$

Both non-overlapped (SLANT-8) and overlapped (SLANT-27) network tiles have been introduced in Figure 2. Briefly, SLANT-8 covered the entire MNI space using eight U-Nets by covering $k = 2 \times 2 \times 2 = 8$ non-overlapped subspaces. Each sub-space in SLANT-8 covered $86 \times 110 \times 78$ voxels (each dimension is about the half the MNI 305 space) with 1mm isotropic resolution. On the other hand, SLANT-27 covered $k = 3 \times 3 \times 3 = 27$ overlapped subspaces. Each sub-space in SLANT-27 covered $96 \times 128 \times 88$ voxels.

*C. Label Fusion*

When separating entire MNI space to $k$ subspaces, the overlapped strategy (as SLANT-27) would provide more than one segmentation results for a single voxel. Herein, an extra step other than concatenation is required to obtain the final segmentation label for that voxel from multiple candidates. In this work, the majority vote label fusion method was employed to get the final segmentation results. Briefly, the majority vote label fusion method was used to fuse $k$ segmentations $\{S_1, S_2, ..., S_k\}$ from network tiles to a single final segmentation $S_{\text{MNI}}$ in MNI space.

$$S_{\text{MNI}}(i) = \underset{l \in \{0,1,...,L-1\}}{\text{argmax}} \frac{1}{k} \sum_{m=1}^{k} p(l|S_m, i) \quad (2)$$

where $\{0,1,...,L-1\}$ represents $L$ possible labels for a given voxel $i$ ($i \in \{1,2,...,N\}$). $p(l|S_m, i) = 1$ if $S_m(i) = l$, and 0, otherwise. The space outside each network tile was excluded in the label fusion. Then, final segmentation in the original target image space was achieved by registering $S_{\text{MNI}}$ to original space from affine registration [11]. Note that for the non-overlapped strategy (as SLANT-8), the naïve concatenation was employed to obtain a final single segmentation directly without using label fusion. If more than one label were equally voted after majority vote for a voxel, the smaller label is used as final label for such voxel.

*D. Boost Learning on Unlabeled Data*

Inspired by [47], we trained the network tile using large-scale auxiliary labels from existing segmentation tools on initially unlabeled MRI scans. In this study, non-local spatial staple label fusion (NLSS) based multi-atlas segmentation pipeline [56] was performed on 5111 multi-site scans. 45 T1-weighted (T1w) MRI scans from Open Access Series on Imaging Studies (OASIS) dataset [57] with BrainCOLOR labeling protocol [58] were used as the atlases. All testing scans and atlases were affinely registered to the MNI305 template [54] using NiftyReg [11]. Before performing the more time-consuming deformable registration, atlas-selection [20, 59-61] is typically performed to select a subset of the most representative atlases to reduce the computational complexity. Practically, 10-20 atlases are sufficient for a good multi-atlas segmentation [20].

In our pipeline, 15 atlases were selected for each testing images by performing a PCA based atlas-selection [36]. To obtain the PCA manifold from all 45 atlases (or 5111+45 atlases for large-scale MAS), the 3D intensities within the same MNI brain mask of each atlas were converted to a 1D vector. Then, a naïve PCA project was performed on 1D vectors from all atlases to learn the PCA manifold. Once a testing scan is projected to the same PCA manifold, the 15 atlases with smallest Euclidean distance to the testing scan were selected as the atlases to segment the testing scans. The 15 selected atlases were pairwise registered [10, 11] and fused to achieve the target segmentation. For non-rigid registration, we use symmetric image normalization (SyN), with a cross correlation similarity metric convergence threshold of $10^{-9}$ and convergence window size of 15, provided by the Advanced Normalization Tools (ANTs) software [10]. For NLSS, the patch neighborhood was set to 3 mm isotropic and the search neighborhood was set to 5 mm isotropic. The spatial standard deviation was set to 1.5 mm. The maximum iteration number was set to 100.

First, the large-scale auxiliary labels were employed to train each network tile, which can be regarded as a pre-training stage. Then, a smaller set of manually traced training images was used to further fine-tune the network tiles. During the fine-tuning, the entire 3D U-Net in network tile was trained without freezing any layers.

## III. CONTAINERIZED IMPLEMENTATION

Since the proposed SLANT method integrates a variety of image processing algorithms (e.g., registration, deep segmentation, label fusion etc.), it might be time-consuming for researchers to repeat the processing outside of our lab. Therefore, we developed a containerized implementation that provided end-to-end segmentation solution using Docker container. Using such implementation, the SLANT method can be deployed on any MRI T1w scans using one line of command.

### A. Whole Brain Segmentation Docker

Docker (https://www.docker.com) is an open-source container technology, which provides a lightweight solution to deploy image processing algorithms in an operating system (OS) independent fashion. Unlike the traditional virtual machine that employs hypervisors to emulate the virtual hardware, the Docker container rests upon a single OS that results in a neater and smaller capsule. To enable the GPU acceleration required by the SLANT method, the NVIDIA-Docker (https://github.com/NVIDIA/nvidia-docker) was employed, which an extension of Docker with GPU capability.

In this paper, we present the SLANT Docker with details of implementation. First, a preprocessing step including N4 bias field correction, intensity normalization, and an affine registration to MNI space has been converted to a single binary executable file using MATLAB mcc complier. Then, the trained network tiles including the Python source code and parameters were included in the Docker. Next, the label fusion and inverse registration to original space has been converted to a single binary executable file using MATLAB mcc complier. The Docker was established on Ubuntu 16.04 with CUDA 8.0, MATLAB 2016a, Python 2.7, and PyTorch 0.2. The Docker has been saved in our dockerhub (https://hub.docker.com/u/masidocker), which can be deployed to the local computer by calling the following command:

*sudo docker pull masidocker/spiders:deep_brain_seg_v1_0_0*

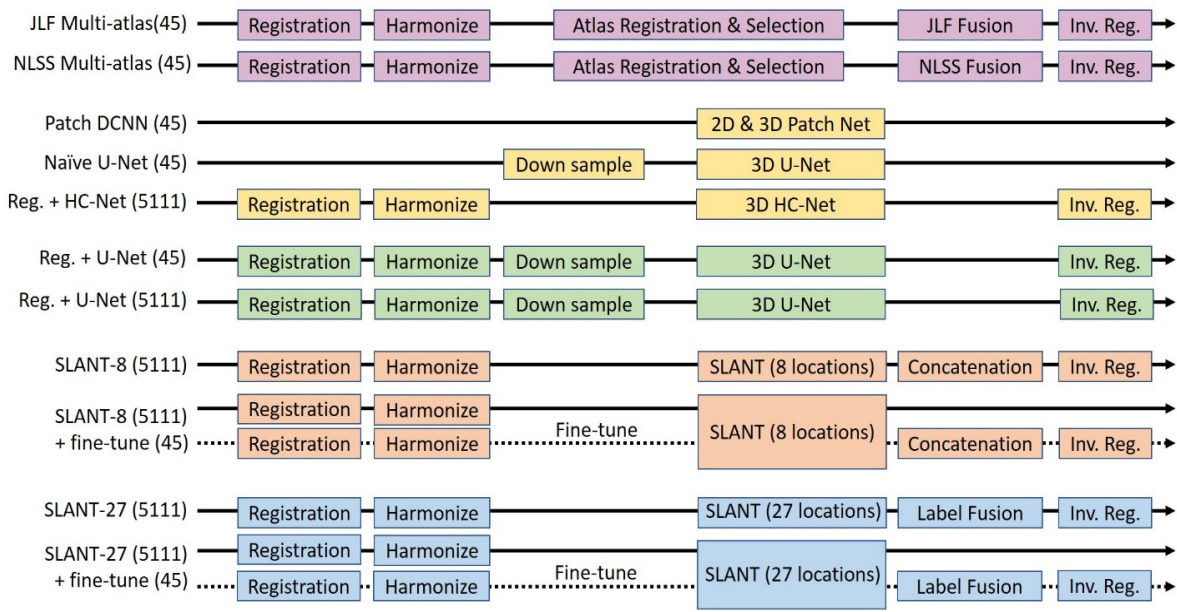

Figure 3. This figure demonstrates the major components for different segmentation methods. "(45)" indicated the 45 OASIS manually traced images were used in training, while "(5111)" indicated the 5111 auxiliary label images were used in training. The joint label fusion (JLF) and non-local spatial STAPLE (NLSS) methods were used as baseline methods.

Table 1. Data summary of 5111 multi-site images.

| Study Name | Website | Images | Sites |
|---|---|---|---|
| Baltimore Longitudinal Study of Aging (BLSA) | www.blsa.nih.gov | 605 | 4 |
| Cutting Pediatrics | vkc.mc.vanderbilt.edu/ebrl | 586 | 2 |
| Autism Brain Imaging Data Exchange (ABIDE) | fcon_1000.projects.nitrc.org/indi/abide | 563 | 17 |
| Information eXtraction from Images (IXI) | www.nitrc.org/projects/ixi_dataset | 523 | 3 |
| Attention Deficit Hyperactivity Disorder (ADHD200) | fcon_1000.projects.nitrc.org/indi/adhd200 | 949 | 8 |
| National Database for Autism Research (NDAR) | ndar.nih.gov | 328 | 6 |
| Open Access Series on Imaging Study (OASIS) | www.oasis-brains.org | 312 | 1 |
| 1000 Functional Connectome (fcon_1000) | fcon_1000.projects.nitrc.org | 1102 | 22 |
| Nathan Kline Institute Rockland (NKI_rockland) | fcon_1000.projects.nitrc.org/indi/enhanced | 143 | 1 |

*B. Run SLANT in Docker*

Once the docker has been imported to the local computer. The users are able to obtain the final output files by running a single command:

*sudo nvidia-docker run -it --rm -v {input path}:/INPUTS/ -v {output path}:/OUTPUTS*
 *masidocker/spiders:deep_brain_seg_v1_0_0 /extra/run_deep_brain_seg.sh*

IV. DATA

One training dataset and three testing datasets have been used in the validation. The training and testing cohort are all MRI T1w 3D volumes. The training strategies of different methods have been shown in Figure 3.

*A. Training Cohort*

The training cohort consisted of 45 T1-weighted (T1w) MRI scans from Open Access Series on Imaging Studies (OASIS) dataset [57]. Each scan was manually traced to 133 labels based on BrainCOLOR protocol [58] by Neuromorphometrics Inc. (http://www.neuromorphometrics.com/). 5111 multi-site T1w MRI scans from nine different projects (Table 1) are used to obtain the large-scale auxiliary training data.

*B. Validation and Testing Cohort*

Three cohorts have been included in this study as validation and testing cohorts. The validation cohort consisted of five T1w MRI scans from the same OASIS dataset as training data. Such dataset was used to decide hyperparameters for training and evaluated the performance of the proposed method on the same site testing data. Then, the remaining testing cohorts were used as independent testing data to evaluate the performance of the proposed method as external validations.

 **OASIS Dataset.** Five withheld T1w MRI scans from OASIS dataset with manual segmentation (BrainCOLOR protocol) were used as the first validation dataset. The resolution of raw T1w scans varies from $256 \times 270 \times 256$ to $256 \times 334 \times 256$, all with 1mm isotropic spatial resolution. The OASIS IDs of the 45 training and five validation scans can be found in (https://github.com/MASILab/SLANTbrainSeg).

**Colin27 Dataset.** The Colin27 T1w MRI scan, a high-quality averaging image from 27 scans from the same subject, was used as the first testing dataset. The high-resolution Colin27 T1w scan [62] was manually traced following BrainCOLOR protocol, which has 0.5mm isotropic spatial resolution with $362 \times 434 \times 362$ voxels. This cohort contained one testing scan, which evaluated the performance of different methods on high-quality and high-resolution scenarios.

**CANDI Dataset.** The second testing cohort contains 13 T1w MRI scans from the Child and Adolescent Neuro Development Initiative (CANDI) [63], which were manually traced following BrainCOLOR protocol. The scans had the same 256x128x256 resolution with 0.94x1.5x0.94 mm voxel size. This testing cohort evaluates the performance of the proposed method on (1) independent external validation dataset, (2) different population, whose age range (5-15 yrs.) was not covered by OASIS training cohort (18-96 yrs.).

V. EXPERIMENTAL DESIGN

*A. Training*

The experimental design for evaluating different methods have been shown in Figure 3. This figure demonstrated the major components for different segmentation methods. "(45)" indicated the 45 OASIS manually traced images were used in training, while "(5111)" indicated the 5111 auxiliary label images were used in training. First, joint label fusion (JLF) [30] and non-local spatial staple (NLSS) [56], two state-of-the-art multi-atlas label fusion methods, were employed as the baseline methods on whole brain segmentation. The hyper-parameters were defined as the recommended values of whole brain segmentation from the publications. The baseline multi-atlas methods used 45 OASIS training data and atlases.

Next, three previously proposed deep learning based whole brain segmentation methods: patch-based network [44], naive 3D U-Net [42], and HC-Net [52] methods were employed as additional CNN baseline methods. Using the affine registration ("Reg") as preprocessing, the "Reg.+U-Net" was also evaluated. To train the "Naïve U-Net", the original T1 MRI scans were resampled to the $96 \times 128 \times 88$ resolution volumes for training using bilinear interpolation. To train the "Reg.+U-Net", the registered T1 MRI scans in MNI space were resampled to the $96 \times 128 \times 88$ resolution volumes for training using bilinear interpolation for intensity scans and nearest neighbor interpolation for label scans. In testing stage, the output segmentation volumes were resampled (and inverse registered for the "Reg.+U-Net") back to the original image space. Such resolution is a compromise considering GPU memory limitation, ratios among three dimensions, and the network design. Both "Naïve U-Net" and "Reg.+U-Net" used only one 3D U-Net shape network. Using 8 or 27 independent U-Net shape networks, the proposed SLANT segmentation pipelines were evaluated on both non-overlapped scenarios ("SLANT-8") and overlapped scenarios ("SLANT-27").

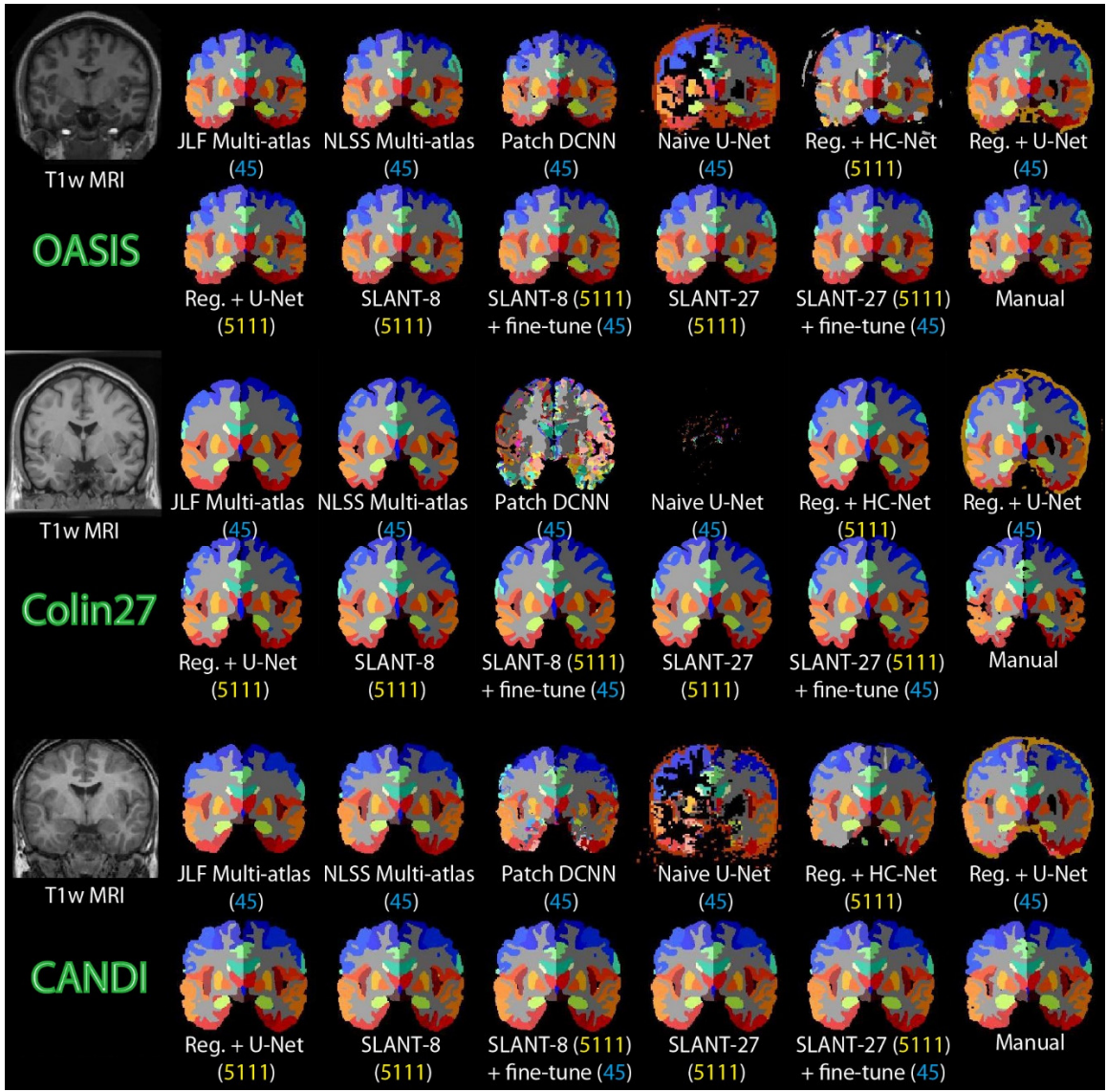

Figure 4. Qualitative results of manual segmentation, multi-atlas segmentation methods, patch based DCNN method, HC-Net, U-Net approaches, and proposed SLANT methods.

The source code of patch-based network [44] and HC-Net [52] were obtained from the link provided by the publications (https://github.com/adbrebs/brain_segmentation and https://github.com/gift-surg/HighRes3DNet), whose hyper-parameters had been kept the same as the publications. For the hyper-parameters that could not be founded in the publications, they were set as the default values in the source code. In Figure 3, "45" means manually labeled scans were used during training, while "5111" means the auxiliary labeled scans were used as training data. The "Reg." means the affine registration has been employed as a preprocessing stage.

To be a fairer comparison, the same 3D segmentation network with the same hyper-parameters has been used for different U-Net (as the entire network) and SLANT experiments (as one network tile among 8 or 27 network tiles). Briefly, batch size = 1, input resolution = $96 \times 128 \times 88$, input channel = 1, output channel = 133, optimizer = "Adam", learning rate = 0.0001. Meanwhile, all the preprocessing and registration methods are kept same for different U-Net and SLANT experiments. All the experimental networks can fit into an NVIDIA Titan GPU with 12 GB memory. For training using 5111 auxiliary label scans, 6 epochs were trained for a 3D segmentation network that each epoch took ~4 training hours. Therefore, training SLANT-27 using 5111 scans on 6 epochs took $27 \times 6 \times 4 = 648$ hours (27 days) on a single GPU. Each network in SLANT-8 or SLANT-27 took more than 11 GB memory on NVIDIA Titan GPU. The detailed computational time of preparing 5111 auxiliary labels, training one epoch, and testing one scan are provided in Table 7. For training from scratch using 45 manual labeled scans, 1000 epochs were trained. For fine-tuning using 45 manual labeled scans, 30 epochs were trained. The results reported in this paper were from the epoch number with the best performance for each method on five OASIS validation images. Then, the network parameters with such epochs were used on testing cohorts Colin27 and CANDI.

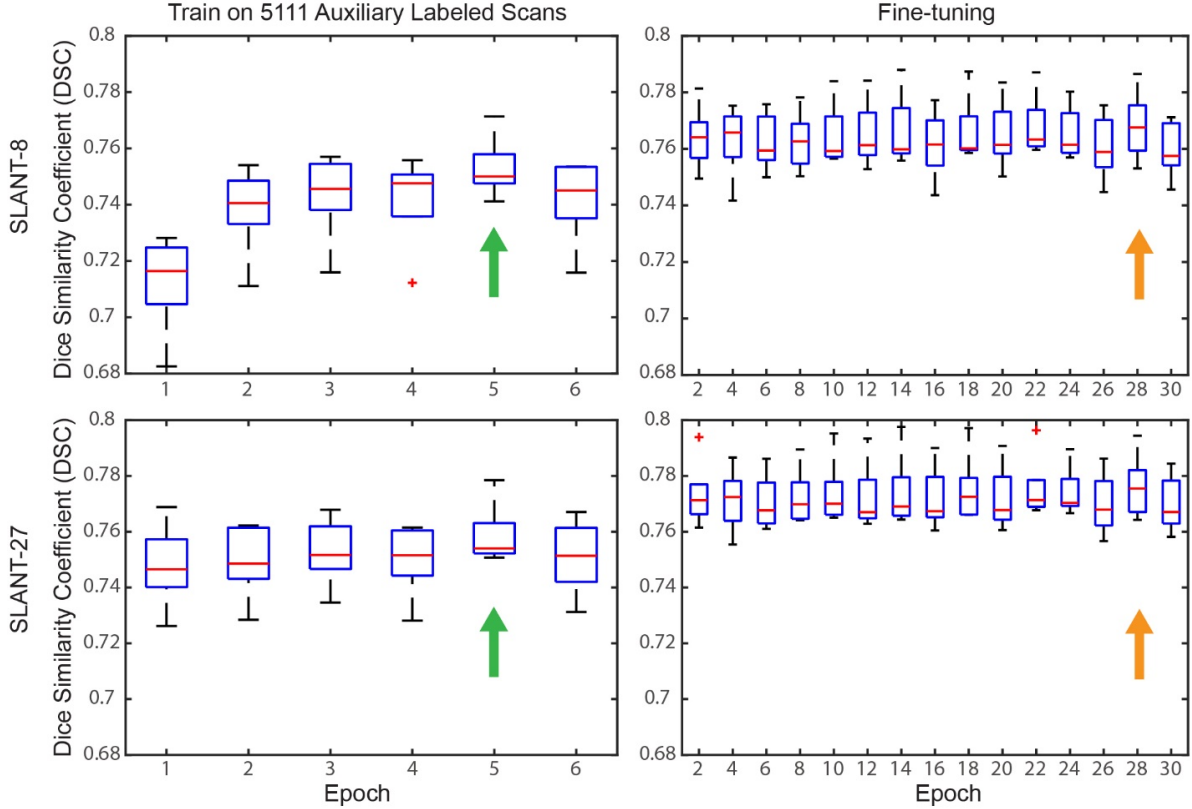

Figure 5. Sensitivity results of training SLANT-8 and SLANT-27. The mean Dice similarity coefficient (DSC) between automatic methods and manual segmentations for different training epochs were showed as boxplots. The Left panels showed the segmentation performance on five OASIS validation cohort using 5111 auxiliary labeled scans. The best performance was from epoch 5, which was used as initial parameters for fine-tuning, whose performance was showed in the right panels. As a result, the model at epoch 28 after fine-tuning was used for SLANT-8 and SLANT-27.

### B. Evaluation Metrics

We employed the Dice similarity coefficients (DSC) as the main evaluation measurement for different approaches by comparing their segmentation results against the ground truth voxel-by-voxel. DSC is a ratio of twice the amount of intersection to the total number of voxels in automatic segmentation $A$ and manual segmentation $M$, which is defined as:

$$DSC = \frac{2|A \cap M|}{|A|+|M|} = \frac{2|TP|}{2|TP|+|FP|+|FN|} \quad (3)$$

where $TP$ is true positive, $FP$ is false positive, $FN$ is false a negative.

Surface error measurements are the complimentary metrics to evaluate the quality of the segmentations. Therefore, we defined the vertices on the automatic segmentation and manual segmentation as $X$ and $Y$ respectively. Then, the mean surface distance (MSD) from the automatic segmentation to manual segmentation is defined as:

$$MSD(X,Y) = \underset{y \in Y}{\mathrm{avg}} \underset{x \in X}{\inf} d(X,Y) \quad (4)$$

where $inf$ represents the infimum, and avg means the average.

The differences between methods were evaluated by Wilcoxon signed rank test [64] and the difference was significant means p<0.05 in this paper.

## VI. RESULTS

### A. Validation

Qualitative results of segmentation from three scans from OASIS validation cohort were shown in Figure 4. The sensitivity analyses had been demonstrated in Figure 5, which presented the model used for validation and testing among different training epochs. The overall segmentation performance on the entire OASIS validation cohort had been shown in Figure 6. In the validation, JLF and NLSS were evaluated as benchmarks using 45 atlases and 5111+45 atlases. The "Patch-DCNN" and "Naïve U-Net" were performed using 45 atlases in the original image space. By introducing the affine registration ("Reg.") as preprocessing, the "Reg. + HC-Net" and "Reg. + U-Net" were conducted as additional benchmarks. The proposed SLANT-8 and SLANT-27 methods were evaluated on using 45 manual atlases, 5111 auxiliary atlases, and 5111+45 atlases.

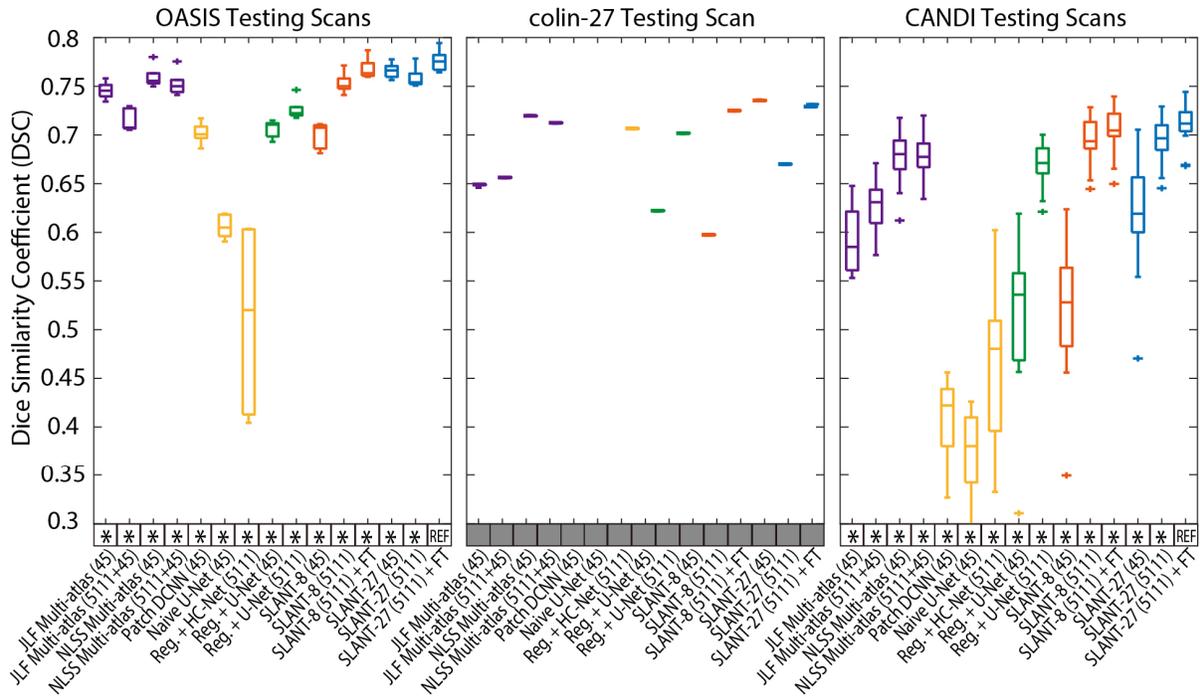

Figure 6. Quantitative results of baseline methods and proposed SLANT methods. The mean Dice similarity coefficient (DSC) between automatic methods and manual segmentations for all testing subjects were showed as boxplots. The SLANT-27 using 5111 auxiliary labels for pretraining and fine-tuned ("FT") by 45 manual labels achieved highest median DSC values and was used as reference method ("REF") in statistical analysis. If the difference to REF was significant from Wilcoxon signed test, the method was marked with "*" symbol.

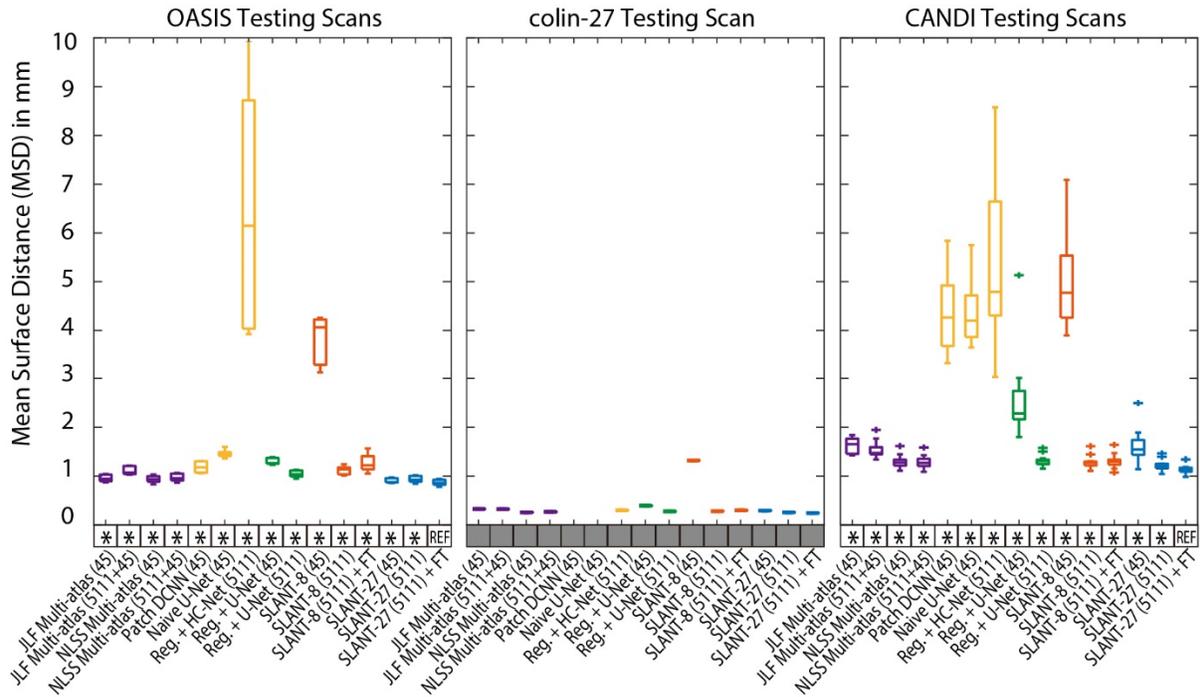

Figure 7. Quantitative results of baseline methods and proposed SLANT methods. The mean surface distance (MSD) between automatic methods and manual segmentations for all testing subjects were showed as boxplots. The SLANT-27 using 5111 auxiliary labels for pretraining and fine-tuned ("FT") by 45 manual labels achieved lowest median MSD values and was used as reference method ("REF") in statistical analysis. If the difference to REF was significant from Wilcoxon signed rank test, the method was marked with "*" symbol.

Table 2. Mean and median Dice similarity coefficients (DSC) values on three validation cohorts

| Methods | Training Scans # | OASIS Dataset | | Colin 27 | CANDI Dataset | |
|---|---|---|---|---|---|---|
| | | mean±std DSC | median DSC | DSC | mean±std DSC | Median DSC |
| JLF | 45 | 0.746±0.009 | 0.746 | 0.646 | 0.590±0.033 | 0.585 |
| JLF | 5111+45 | 0.715±0.012 | 0.708 | 0.653 | 0.626±0.029 | 0.631 |
| NLSS | 45 | 0.760±0.012 | 0.756 | 0.712 | 0.677±0.029 | 0.680 |
| NLSS | 5111+45 | 0.752±0.013 | 0.750 | 0.705 | 0.680±0.025 | 0.677 |
| Patch DCNN | 45 | 0.702±0.011 | 0.701 | 0.012 | 0.409±0.038 | 0.422 |
| Naive U-Net | 45 | 0.606±0.012 | 0.605 | 0.000 | 0.375±0.043 | 0.380 |
| Reg. + HC-Net | 5111 | 0.509±0.097 | 0.520 | 0.700 | 0.434±0.146 | 0.480 |
| Reg. + U-Net | 45 | 0.706±0.009 | 0.711 | 0.621 | 0.514±0.081 | 0.536 |
| Reg, + U-Net | 5111 | 0.726±0.012 | 0.722 | 0.695 | 0.669±0.023 | 0.671 |
| SLANT-8 | 45 | 0.699±0.014 | 0.707 | 0.597 | 0.519±0.070 | 0.528 |
| SLANT-8 | 5111 | 0.753±0.011 | 0.750 | 0.717 | 0.694±0.024 | 0.694 |
| SLANT-8 + FT | 5111+45 | 0.768±0.011 | 0.763 | 0.726 | 0.704±0.025 | 0.705 |
| SLANT-27 | 45 | 0.766±0.008 | 0.766 | 0.665 | 0.621±0.062 | 0.619 |
| SLANT-27 | 5111 | 0.759±0.011 | 0.754 | 0.721 | 0.694±0.024 | 0.697 |
| SLANT-27 + FT | 5111+45 | **0.776**±0.012 | **0.775** | **0.732** | **0.711**±0.023 | **0.712** |

Table 3. Mean and median mean surface distance (MSD) values (mm) on three validation cohorts

| Methods | Training Scans # | OASIS Dataset | | Colin 27 | CANDI Dataset | |
|---|---|---|---|---|---|---|
| | | mean±std MSD | median MSD | MSD | mean±std MSD | Median MSD |
| JLF | 45 | 0.956±0.075 | 0.943 | 0.328 | 1.635±0.152 | 1.654 |
| JLF | 5111+45 | 1.112±0.090 | 1.071 | 0.325 | 1.540±0.167 | 1.475 |
| NLSS | 45 | 0.938±0.076 | 0.935 | 0.260 | 1.289±0.136 | 1.280 |
| NLSS | 5111+45 | 0.972±0.084 | 0.955 | 0.268 | 1.285±0.131 | 1.272 |
| Patch DCNN | 45 | 1.185±0.120 | 1.180 | 10.494 | 4.392±0.789 | 4.267 |
| Naive U-Net | 45 | 1.458±0.085 | 1.448 | 11.673 | 4.377±0.677 | 4.201 |
| Reg. + HC-Net | 5111 | 6.478±2.630 | 6.148 | 0.301 | 6.591±5.242 | 4.793 |
| Reg. + U-Net | 45 | 1.302±0.068 | 1.267 | 0.394 | 2.548±0.853 | 2.281 |
| Reg, + U-Net | 5111 | 1.046±0.075 | 1.038 | 0.277 | 1.313±0.118 | 1.306 |
| SLANT-8 | 45 | 3.798±0.534 | 4.065 | 1.320 | 4.999±0.898 | 4.774 |
| SLANT-8 | 5111 | 1.118±0.091 | 1.138 | 0.285 | 1.284±0.127 | 1.274 |
| SLANT-8 + FT | 5111+45 | 1.273±0.196 | 1.221 | 0.299 | 1.303±0.139 | 1.296 |
| SLANT-27 | 45 | 0.907±0.051 | 0.878 | 0.297 | 1.606±0.349 | 1.544 |
| SLANT-27 | 5111 | 0.938±0.071 | 0.923 | 0.259 | 1.219±0.112 | 1.209 |
| SLANT-27 + FT | 5111+45 | **0.867**±0.065 | **0.860** | **0.246** | **1.141**±0.106 | **1.125** |

Table 4. Mean and median Hausdorff distance (HD) values (mm) on three validation cohorts

| Methods | Training Scans # | OASIS Dataset | | Colin 27 | CANDI Dataset | |
|---|---|---|---|---|---|---|
| | | mean±std HD | median HD | HD | mean±std HD | Median HD |
| JLF | 45 | 8.161±0.572 | 8.252 | 2.786 | 10.167±0.770 | 9.941 |
| JLF | 5111+45 | 8.524±0.512 | 8.484 | 2.754 | 9.572±0.705 | 9.414 |
| NLSS | 45 | **7.926**±0.527 | **7.935** | **2.537** | 9.066±0.592 | 9.042 |
| NLSS | 5111+45 | 8.031±0.493 | 7.954 | 2.572 | **9.063**±0.649 | 8.939 |
| Patch DCNN | 45 | 28.501±11.695 | 25.113 | 27.111 | 40.425±3.533 | 41.306 |
| Naive U-Net | 45 | 18.454±5.781 | 16.837 | 18.733 | 60.815±4.595 | 60.764 |
| Reg. + HC-Net | 5111 | 76.466±6.160 | 77.810 | 4.263 | 64.171±6.088 | 64.654 |
| Reg. + U-Net | 45 | 13.178±2.450 | 13.072 | 6.030 | 38.584±4.926 | 38.816 |
| Reg, + U-Net | 5111 | 8.336±0.551 | 8.119 | 2.590 | 9.538±0.933 | 9.364 |
| SLANT-8 | 45 | 44.058±3.895 | 44.844 | 13.353 | 54.789±2.384 | 56.146 |
| SLANT-8 | 5111 | 30.791±5.030 | 30.037 | 7.167 | 24.196±3.572 | 24.321 |
| SLANT-8 + FT | 5111+45 | 33.670±6.442 | 32.535 | 7.700 | 28.605±3.712 | 29.743 |
| SLANT-27 | 45 | 9.993±1.165 | 9.867 | 3.913 | 20.901±4.613 | 20.648 |
| SLANT-27 | 5111 | 8.247±0.818 | 8.002 | 2.578 | 9.195±1.003 | 9.083 |
| SLANT-27 + FT | 5111+45 | 8.224±0.775 | 8.267 | 2.643 | 9.328±1.273 | **8.901** |

Table 5. List of ROIs

| | | | |
|---|---|---|---|
| 0 | Unlabeled | 138 | Right MCgG   middle cingulate gyrus |
| 4 | 3rd Ventricle | 139 | Left MCgG   middle cingulate gyrus |
| 11 | 4th Ventricle | 140 | Right MFC   medial frontal cortex |
| 23 | Right Accumbens Area | 141 | Left MFC   medial frontal cortex |
| 30 | Left Accumbens Area | 142 | Right MFG   middle frontal gyrus |
| 31 | Right Amygdala | 143 | Left MFG   middle frontal gyrus |
| 32 | Left Amygdala | 144 | Right MOG   middle occipital gyrus |
| 35 | Brain Stem | 145 | Left MOG   middle occipital gyrus |
| 36 | Right Caudate | 146 | Right MOrG   medial orbital gyrus |
| 37 | Left Caudate | 147 | Left MOrG   medial orbital gyrus |
| 38 | Right Cerebellum Exterior | 148 | Right MPoG   postcentral gyrus medial segment |
| 39 | Left Cerebellum Exterior | 149 | Left MPoG   postcentral gyrus medial segment |
| 40 | Right Cerebellum White Matter | 150 | Right MPrG   precentral gyrus medial segment |
| 41 | Left Cerebellum White Matter | 151 | Left MPrG   precentral gyrus medial segment |
| 44 | Right Cerebral White Matter | 152 | Right MSFG   superior frontal gyrus medial segment |
| 45 | Left Cerebral White Matter | 153 | Left MSFG   superior frontal gyrus medial segment |
| 47 | Right Hippocampus | 154 | Right MTG   middle temporal gyrus |
| 48 | Left Hippocampus | 155 | Left MTG   middle temporal gyrus |
| 49 | Right Inf Lat Vent | 156 | Right OCP   occipital pole |
| 50 | Left Inf Lat Vent | 157 | Left OCP   occipital pole |
| 51 | Right Lateral Ventricle | 160 | Right OFuG   occipital fusiform gyrus |
| 52 | Left Lateral Ventricle | 161 | Left OFuG   occipital fusiform gyrus |
| 55 | Right Pallidum | 162 | Right OpIFG opercular part of the inferior frontal gyrus |
| 56 | Left Pallidum | 163 | Left OpIFG opercular part of the inferior frontal gyrus |
| 57 | Right Putamen | 164 | Right OrIFG orbital part of the inferior frontal gyrus |
| 58 | Left Putamen | 165 | Left OrIFG orbital part of the inferior frontal gyrus |
| 59 | Right Thalamus Proper | 166 | Right PCgG   posterior cingulate gyrus |
| 60 | Left Thalamus Proper | 167 | Left PCgG   posterior cingulate gyrus |
| 61 | Right Ventral DC | 168 | Right PCu   precuneus |
| 62 | Left Ventral DC | 169 | Left PCu   precuneus |
| 71* | Cerebellar Vermal Lobules I-V | 170 | Right PHG   parahippocampal gyrus |
| 72* | Cerebellar Vermal Lobules VI-VII | 171 | Left PHG   parahippocampal gyrus |
| 73* | Cerebellar Vermal Lobules VIII-X | 172 | Right PIns   posterior insula |
| 75 | Left Basal Forebrain | 173 | Left PIns   posterior insula |
| 76 | Right Basal Forebrain | 174 | Right PO   parietal operculum |
| 100 | Right ACgG   anterior cingulate gyrus | 175 | Left PO   parietal operculum |
| 101 | Left ACgG   anterior cingulate gyrus | 176 | Right PoG   postcentral gyrus |
| 102 | Right AIns   anterior insula | 177 | Left PoG   postcentral gyrus |
| 103 | Left AIns   anterior insula | 178 | Right POrG   posterior orbital gyrus |
| 104 | Right AOrG   anterior orbital gyrus | 179 | Left POrG   posterior orbital gyrus |
| 105 | Left AOrG   anterior orbital gyrus | 180 | Right PP   planum polare |
| 106 | Right AnG   angular gyrus | 181 | Left PP   planum polare |
| 107 | Left AnG   angular gyrus | 182 | Right PrG   precentral gyrus |
| 108 | Right Calc   calcarine cortex | 183 | Left PrG   precentral gyrus |
| 109 | Left Calc   calcarine cortex | 184 | Right PT   planum temporale |
| 112 | Right CO   central operculum | 185 | Left PT   planum temporale |
| 113 | Left CO   central operculum | 186 | Right SCA   subcallosal area |
| 114 | Right Cun   cuneus | 187 | Left SCA   subcallosal area |
| 115 | Left Cun   cuneus | 190 | Right SFG   superior frontal gyrus |
| 116 | Right Ent   entorhinal area | 191 | Left SFG   superior frontal gyrus |
| 117 | Left Ent   entorhinal area | 192 | Right SMC   supplementary motor cortex |
| 118 | Right FO   frontal operculum | 193 | Left SMC   supplementary motor cortex |
| 119 | Left FO   frontal operculum | 194 | Right SMG   supramarginal gyrus |
| 120 | Right FRP   frontal pole | 195 | Left SMG   supramarginal gyrus |
| 121 | Left FRP   frontal pole | 196 | Right SOG   superior occipital gyrus |
| 122 | Right FuG   fusiform gyrus | 197 | Left SOG   superior occipital gyrus |
| 123 | Left FuG   fusiform gyrus | 198 | Right SPL   superior parietal lobule |
| 124 | Right GRe   gyrus rectus | 199 | Left SPL   superior parietal lobule |
| 125 | Left GRe   gyrus rectus | 200 | Right STG   superior temporal gyrus |
| 128 | Right IOG   inferior occipital gyrus | 201 | Left STG   superior temporal gyrus |
| 129 | Left IOG   inferior occipital gyrus | 202 | Right TMP   temporal pole |
| 132 | Right ITG   inferior temporal gyrus | 203 | Left TMP   temporal pole |
| 133 | Left ITG   inferior temporal gyrus | 204 | Right TrIFG triangular part of the inferior frontal gyrus |
| 134 | Right LiG   lingual gyrus | 205 | Left TrIFG triangular part of the inferior frontal gyrus |
| 135 | Left LiG   lingual gyrus | 206 | Right TTG   transverse temporal gyrus |
| 136 | Right LOrG   lateral orbital gyrus | 207 | Left TTG   transverse temporal gyrus |
| 137 | Left LOrG   lateral orbital gyrus | | |

"*" indicates the three labels that were not included in Colin27 and CANDI cohorts.

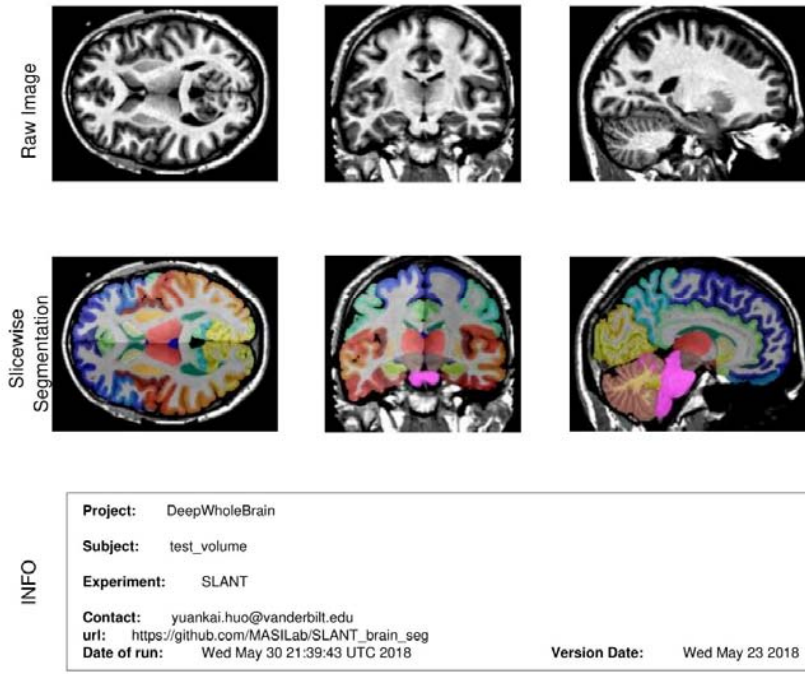

Figure 8. The screenshot of Docker output report designed by us. The users will able to review the segmentation quality immediately after the scan.

Table 2. Number of Best Performance among ROIs

| Methods | Δ=0 | Δ=0.01 | Δ=0.02 | Δ=0.03 | Δ=0.04 | Δ=0.05 |
|---|---|---|---|---|---|---|
| JLF | 1 | 1 | 2 | 4 | 7 | 11 |
| NLSS | 1 | 10 | 34 | 56 | 83 | 99 |
| Patch DCNN | 1 | 1 | 1 | 1 | 1 | 3 |
| Reg, + U-Net (5111) | 1 | 2 | 11 | 29 | 53 | 78 |
| SLANT-8 + FT | 40 | 79 | 105 | 116 | 122 | 123 |
| SLANT-27 + FT | **85** | **102** | **113** | **118** | **124** | **128** |

"Δ" means the range of tolerance to be the best performance algorithm.

Table 3. Summary of Computational Time

| Methods | Training Scans # | Auxiliary Labels Generation Time using One CPU or GPU | Training Time For One Epoch | Testing Time on One Scan |
|---|---|---|---|---|
| JLF | 45 | N/A | N/A | ≈ 34 hours |
| JLF | 5111+45 | ≈ 21 CPU years **or** 2 GPU months | N/A | ≈ 34 hours |
| NLSS | 45 | N/A | N/A | ≈ 36 hours |
| NLSS | 5111+45 | ≈ 21 CPU years **or** 2 GPU months | N/A | ≈ 36 hours |
| Patch DCNN | 45 | N/A | ≈ 2 mins | ≈ 1 min |
| Naive U-Net | 45 | N/A | ≈ 2 mins | ≈ 1 min |
| Reg. + HC-Net | 5111 | ≈ 21 CPU years **or** 2 GPU months | ≈ 4 h | ≈ 8 mins |
| Reg. + U-Net | 45 | N/A | ≈ 2 mins | ≈ 8 mins |
| Reg, + U-Net | 5111 | ≈ 21 CPU years **or** 2 GPU months | ≈ 4 h (**or** 2 mins with 27 GPUs) | ≈ 8 mins |
| SLANT-8 | 45 | N/A | ≈ 16 mins (**or** 2 mins with 8 GPUs) | ≈ 10 mins |
| SLANT-8 | 5111 | ≈ 21 CPU years **or** 2 GPU months | ≈ 32 h (**or** 4 h with 8 GPUs) | ≈ 10 mins |
| SLANT-8 + FT | 5111+45 | ≈ 21 CPU years **or** 2 GPU months | ≈ 32.25 h (**or** 4 h with 8 GPUs) | ≈ 10 mins |
| SLANT-27 | 45 | N/A | ≈ 1 h (**or** 2 mins with 27 GPUs) | ≈ 15 mins |
| SLANT-27 | 5111 | ≈ 21 CPU years **or** 2 GPU months | ≈ 108 h (**or** 4 h with 27 GPUs) | ≈ 15 mins |
| SLANT-27 + FT | 5111+45 | ≈ 21 CPU years **or** 2 GPU months | ≈ 109 h (**or** 4 h with 27 GPUs) | ≈ 15 mins |

\* The testing times for the deep learning based methods include all required processing time beyond GPU time (starting time, registration, inverse-registration etc.) for one testing scan.

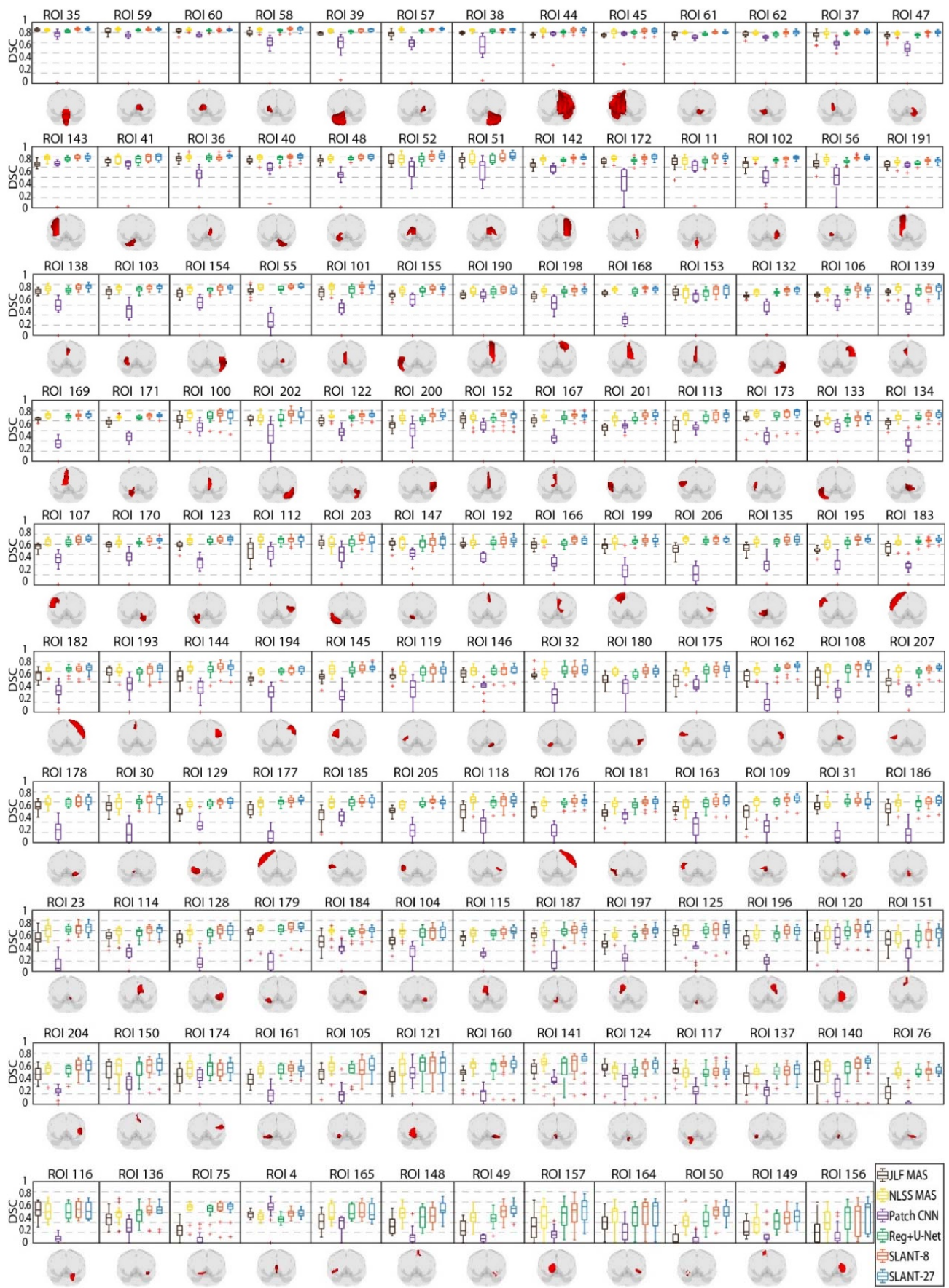

Figure 9. Quantitative results of all regions of interests (ROIs) by comparing SLANT-27 and representative baseline methods.

*B. Testing*

The qualitative performance on Colin27 and CANDI was presented in Figure 4. In Figure 6, the quantitative results of baseline methods and proposed SLANT methods. The mean Dice similarity coefficient (DSC) between automatic methods and manual segmentations was calculated for all testing subjects were showed as boxplots. The SLANT-27 using 5111 auxiliary labels for pretraining and fine-tuned ("FT") by 45 manual labels achieved highest median DSC values and was used as a reference method ("REF") in statistical analysis. If the difference to REF was significant from Wilcoxon signed rank test, the method was marked with "*" symbol. The mean DSC values on all anatomical labels (excluding background) between automated methods and manual tracing in original image space were showed as boxplots. From the results, affine registration ("Reg. + U-Net") leveraged the performance of "U-Net" significantly, compared with "Naïve U-Net". For the same "Reg. + U-Net" network, training strategy using 5111 auxiliary labeled scans achieved superior segmentation performance than the strategy only using 45 manual labels. In Figure 7, the quantitative results were also presented as the mean surface distance (MSD) values, which demonstrated that the proposed SLANT-27 method using 5111 auxiliary labels and fine-tuning achieved superior performance compared with baseline methods. The Table 2 3, and 4 presented the detailed quantitative measurement for all methods in terms of DSC, MSD, and Hausdorff distance, which also posed that the proposed SLANT-27 method with fine-tuning achieved the best performance validation and testing cohorts. As shown in Figure 4, the "Patch-DCNN" and "Naïve U-Net" did not achieve meaningful segmentation results for Colin 27 since it is difficult for the network to capture the complicated spatial variations (e.g., different original spatial locations) from only 45 training scans. However, after reducing such spatial variations by introducing a simple affine-registration, the same U-Net "(Reg+U-Net" with 45 training scans) achieved large improvements on DSC. As the entire proposed pipeline has been implemented in a docker container, we defined our standard report (Figure 8) for quality assurance (QA) purpose. As a result, users will able to review the segmentation results immediately after finishing the segmentation.

The performance of each anatomical region was showed in Figure 9, which was presented as box plots. Totally 129 regions have been compared since the label 71, 72, and 73 were not defined in the manual segmentation of testing cohorts. The brain regions defined in BrainCOLOR protocol was presented in Table 5. To visualize the performance of all regions in an organized manner, we sorted the ROIs based on 1D low dimensional feature values from multidimensional scaling (MDS). Briefly, the mean, median, and std of each method were concatenated to a 2D matrix, which each row is a subject. Then, the MDS was deployed on such a matrix to have a 1D low dimensional representation of results for each brain region. We plotted the results of all brain regions from the smallest MDS value (top left) to largest (bottom right) to visualization the patterns of segmentation performances. From the results, the larger brain regions typically yielded better segmentation performance. The number of brain regions that a method achieved the best median DSC performance was showed in Table 6. The "Δ" indicated the tolerable DSC differences that considered the methods to have equal performance. Each number in Table 6 indicated that the number of overall best performance of such method on all ROIs with a certain value of Δ. From the results, the proposed SLANT-27 method consistently achieved the largest number of best performance regions.

VII. CONCLUSION

In this study, we developed the SLANT high-resolution whole brain segmentation method, which combined the canonical medical image processing approaches (registration, harmonization, label fusion) with the deep neural networks. The proposed method addresses the GPU memory limitation on high-resolution 3D learning by introducing the network tiles. The SLANT network tiles used multiple spatially distributed 3D networks to learn segmentation at different spatial locations rather than learning all 3D patches at different spatial locations using a single network. To achieve decent segmentation performance with limited manually labeled whole brain segmentation scans, a large-scale 5111 initially unlabeled MRI T1w scans were used as auxiliary training data by applying the multi-atlas segmentation.

The internal validation and external validation (testing) have been deployed on the trained models, whose corresponding epoch number was learned from sensitivity analyses (Figure 5). From Table 2 and Table 3, the proposed SLANT-27 method achieved better overall segmentation performance. The final qualitative results were shown in Figure 4, while the quantitative results were shown in Figure 6 and Figure 7 respectively in terms of volume and surface evaluation methods. Moreover, the proposed method requires ~15 minutes, compared with >30 hours are typically required by multi-atlas segmentation methods.

VIII. DISCUSSION

The network tile has been proposed to address the memory issues for high-resolution brain segmentation. In this study, 27 tiles are used for segmenting 1mm isotropic resolution MRI scans. In the future, the memory for a single GPU would be enough for housing the entire scan with the rapid development of hardware. However, the memory limitation would still exist when segmenting higher resolution MRI scans (e.g., 0.5 mm isotropic resolution MRI scans or even histology scans). Therefore, the proposed network tile strategy could be adapted for such applications.

The major limitation of the proposed method is that the larger computational resource is required. If multiple GPUs are available (e.g., eight GPUs for SLANT-8, 27 GPUs for SLANT-27), both training and testing time would be the same as single GPU based methods. Otherwise, when only a single GPU is available, the computational time for both training and testing will be linearly increased with the number of network tiles. The NLSS multi-atlas segmentation pipeline is used as one benchmark method and also for auxiliary training. Around 36 hours are typically required since 15 pairwise registration (~2 hours per pair) and non-local search label fusion (~ 6 hours) are employed. Many efforts have been proposed to accelerate the process using GPU acceleration [65], atlas selection [25], fast patch matching [66] etc. The speed of whole brain segmentation could be accelerated using such techniques.

In this study, MAS using the NLSS is employed to achieve 5111 automatic auxiliary segmentation, which is an accurate but resource intensive segmentation method. For instance, if each scan needs 36 hours using multi-atlas segmentation, 21 computational years are required for a single workstation. Meanwhile, the large-scale medical image processing infrastructure [67] and high performance computing cluster at Vanderbilt University were used. These resources access more than 10,000 computational cores. Therefore, 5111 multi-atlas segmentation jobs can be finished in two days if using about half of the ACCRE's resource (at an expense that would be impractical for routine inquiry). Practically, we typically used about 100 cores, which took more 2.5 month to process all data. Now, since we have made the SLANT Docker container freely available online (https://github.com/MASILab/SLANTbrainSeg), it only takes 15 minutes to segment a single scan. Therefore, a single workstation with a single GPU could process more than 5000 MRI scans within two months on a single work station, which enables other researchers to perform the auxiliary training using their own data. If multiple GPUs are available, the computational time to prepare auxiliary segmentations could be further reduced.

Another limitation is that a traditional affine registration is used, which takes about 5 minutes. In the future, the computational time of such registration could be alleviated using deep learning based affine registration methods. Meanwhile, NVIDIA Titan GPU with 12GB memory was used to train the SLANT network. If the less GPU memory is available, the input size of each patch could be reduced be compatible with the hardware. However, that might lead to even more atlas tiles compared with SLANT-8 and SLANT-27.

In this study, 45 atlases were used as fully annotated training data, which contains the 35 open source atlases from MICCAI 2012 challenge on multi-atlas labelling. In the large number of previous publications (which used the challenge data), 15 were typically used as training while 20 were used as testing. Since 2012, we have acquired more data with manual segmentation. We decided to use 45 atlases rather than 15 atlases for training the SLANT methods. The rationale is to maximize the segmentation performance on behalf of training data since more training data typically yield to superior performance for learning based methods. To compare with the state-of-the-art multi-atlas segmentation methods, the benchmarks in this study were all performed using the same 45 atlases with the recommended hyper-parameters in the publications.

This work demonstrates the value of canonical medical image processing approaches in this deep learning era. Thanks to affine registration, the performance of U-Net was leveraged by a significant margin ("Reg.+U-Net (45)" vs. "Naïve U-Net (45)"). Moreover, the same network achieved superior performance with more training data ("Reg.+U-Net (5111)" vs. "Reg.+U-Net (45)"), even were acquired from other automatic methods rather than manual delineations. Note that, more computational resources are required to train a network using 5111 auxiliary labeled scans compared with only using 45 scans. Therefore, the faster training strategy or adaptive sampling strategies could be the future directions to further accelerate the process.

The Patch DCNN and HC-Net were run based on the default hyperparameters in the code and the optimal parameters in the publications. We used such hyperparameters directly since they were published and have been tuned for brain segmentation tasks. Meanwhile, the same standard 3D U-Net hyper-parameters are used to train the "Naïve U-Net" and "Reg.+U-Net" (on both 45 and 5111 down-sampled scans) without heavy tuning. Then, the same hyper-parameters are directly used to train the proposed SLANT-8 and SLANT-27 without additional tuning. Therefore, the same hyper-parameters are used to train "Naïve U-Net", "Reg.+U-Net", SLANT-8 and SLANT-27 without individual tuning (the hyper-parameters can be found in https://github.com/MASILab/SLANTbrainSeg). When presenting the results, different epoch numbers with optimal performance on validation cohort are selected for the different methods respectively. Then the trained models corresponding to such epoch numbers are directly applied to the independent testing cohorts without further tuning. However, the performance of both benchmark methods and proposed methods could be further improved by performing additional tuning of the hyper-parameters on validation cohort respectively.

In this study, only 2×2×2 (SLANT-8) and 3×3×3 (SLANT-27) configurations are evaluated since they lead to a better balance between performance and computational cost compared with the larger number of tiles (e.g., 4×4×4 (SLANT-64) or more). The SLANT-27 achieved better performance compared with the SLANT-8 by introducing (1) more overlapped spatial locations, and (2) multi-atlas label fusion. In our pipeline, only the majority vote is employed. However, other label fusion methods might further leverage the segmentation performance. Another interesting finding is that when the "SLANT-27 (5111)" was only trained by NLSS labels, it achieved better performance than NLSS on testing dataset. Meanwhile, the 3D U-Net was used as the network tile in this work, which can be replaced by other 3D segmentation networks.

IX. ACKNOWLEDGMENT

This research was supported by NSF CAREER 1452485, NIH grants 5R21EY024036, R01EB017230 (Landman), R01NS095291 (Dawant). This research was conducted with the support from Intramural Research Program, National Institute on Aging, NIH. This work was also supported by the National Institutes of Health in part by the National Institute of Biomedical Imaging and Bioengineering training grant T32-EB021937. This study was in part using the resources of the Advanced Computing Center for Research and Education

(ACCRE) at Vanderbilt University, Nashville, TN. This project was supported in part by ViSE/VICTR VR3029 and the National Center for Research Resources, Grant UL1 RR024975-01, and is now at the National Center for Advancing Translational Sciences, Grant 2 UL1 TR000445-06. We appreciate the NIH S10 Shared Instrumentation Grant 1S10OD020154-01 (Smith), Vanderbilt IDEAS grant (Holly-Bockelmann, Walker, Meliler, Palmeri, Weller), and ACCRE's Big Data TIPs grant from Vanderbilt University. We gratefully acknowledge the support of NVIDIA Corporation with the donation of the Titan X Pascal GPU used for this research.

We would like to thank all of the participants that made this work possible. Additionally, we'd like to thank and acknowledge the open access data platforms and data sources that were used for this work, including: XNAT, Neuroimaging Informatics Tool and Resources Clearinghouse (NITRC), 1000 Functional Connectomes Project (fcon-1000), Autism Brain Imaging Data Exchange (ABIDE I), Attention Deficit Hyperactivity Disorder 200 (ADHD-200), Baltimore Longitudinal Study of Aging (BLSA), Education and Brain sciences Research Lab (EBRL), Information eXtraction from Images (IXI; http://brain-development.org/ixi-dataset), National Database for Autism Research (NDAR; NIH MRI Study of Normal Brain Development, dataset #1151; http://www.bic.mni.mcgill.ca/nihpd/info/participating_centers.html), Nathan Kline Institute Rockland Sample (NKI Rockland), and Open Access Series of Imaging Studies (OASIS). The NIH MRI Study of Normal Brain Development is supported by the National Institute of Child Health and Human Development, the National Institute on Drug Abuse, the National Institute of Mental Health, and the National Institute of Neurological Disorders and Stroke— Contract #s N01-HD02-3343, N01-MH9-0002, and N01-NS-9-2314, N01-NS-9-2315, N01-NS-9-2316, N01-NS-9-2317, N01-NS-9-2319 and N01-NS-9-2320. This manuscript reflects the views of the authors and may not reflect the opinions or views of the NIH. Additional funding sources can be found at: http://fcon_1000.projects.nitrc.org/fcpClassic/FcpTable.html (1000 Functional Connectomes Project); http://fcon_1000.projects.nitrc.org/indi/abide/abide_I.html (ABIDE); http://fcon_1000.projects.nitrc.org/indi/adhd200/ (ADHD-200); http://fcon_1000.projects.nitrc.org/indi/enhanced/ (NKI Rockland); http://www.oasis-brains.org/ (OASIS).